\documentclass{article}

\usepackage{arxiv}

\usepackage[utf8]{inputenc} % allow utf-8 input
\usepackage[T1]{fontenc}    % use 8-bit T1 fonts
\usepackage{hyperref}       % hyperlinks
\usepackage{url}            % simple URL typesetting
\usepackage{booktabs}       % professional-quality tables
\usepackage{amsfonts}       % blackboard math symbols
\usepackage{nicefrac}       % compact symbols for 1/2, etc.
\usepackage{microtype}      % microtypography
\usepackage{lipsum}
\usepackage{graphicx}
\graphicspath{ {./images/} }
\newcommand{\etal}{\textit{et al.}}

\title{Motion Prediction Performance Analysis for Autonomous Driving Systems and the Effects of Tracking Noise}

\author{
 Ameni Trabelsi \\
  Computer Science Department\\
  Colorado State University\\
  Fort Collins, CO 80526 \\
  \texttt{ameni.trabelsi@colostate.edu} \\
   \And
 Ross Beveridge \\
  Computer Science Department\\
  Colorado State University\\
  Fort Collins, CO 80526 \\
  \texttt{Ross.Beveridge@colostate.edu} \\
  \And
 Nathaniel Blanchard \\
  Computer Science Department\\
  Colorado State University\\
  Fort Collins, CO 80526 \\
  \texttt{nathaniel.blanchard@colostate.edu} \\
}

\begin{document}
\maketitle
\begin{abstract}
Autonomous driving consists of a multitude of interacting modules, where each module must contend with errors from the others. Typically, the motion prediction module depends upon a robust tracking system to capture each agent's past movement. In this work, we systematically explore the importance of the tracking module for the motion prediction task and ultimately conclude that the overall motion prediction performance is highly sensitive to the tracking module's imperfections. We explicitly compare models that use tracking information to models that do not across multiple scenarios and conditions. We find that the tracking information plays an essential role and improves motion prediction performance in noise-free conditions. However, in the presence of tracking noise, it can potentially affect the overall performance if not studied thoroughly. We thus argue practitioners should be mindful of noise when developing and testing motion/tracking modules, or that they should consider tracking free alternatives.  
\end{abstract}

% keywords can be removed

\section{Introduction}

Autonomous driving depends upon a mixture of perception modules to achieve safe motion planning.
%involves a mixture of perception modules to achieve safe motion planning. 
Perception must unfold in highly uncertain, rapidly changing, and interactive environments shared with other dynamic agents. Planning focuses on the real-time, safe navigation of such an environment. 
In this work, we focus on the two perceptive tasks of agent tracking and motion prediction. 
Typically, these two tasks are cascaded; agent tracking output feeds into motion prediction.
%the output of agent tracking explicitly inputs into motion prediction. 
Such cascaded approaches are usually highly affected by errors propagating from noisy components.
For instance, errors propagated from a noisy tracking module can hinder the performance of the motion prediction and planning modules. Such problems can result in catastrophic failures as the system fails to recover from errors accumulated through the pipeline. 

%Numerous sources of noise on the input to tracking will result in labeling errors in the output. 
Despite the complexities of cascaded interactions, most works on these topics do not examine how errors propagate and affect downstream modules. In this work, we ask a novel question: does the tracking system, a sub-component of the motion prediction, contribute to overall accuracy improvements in real-world settings. We focus on the tracking module due to the propensity  of noise in real-world environments, a reality of several common autonomous driving issues like heavy occlusion, crowded scenes, high inter-frame motion, and camera motion. Thus, we study the effect of tracking noise on the motion prediction performance and compare the tracking-based to the tracking-free alternatives under various conditions. We show that, effective and robust trackers introducing little to no noise can play an essential role in the performance improvement of motion prediction models that use the tracking information. However, practitioners should be aware of the effect of tracking noise on the motion prediction performance and should consider that when selecting their approach. Finally, the tracking-free models can be, in the case of an inevitable noisy tracker, a potential alternative that is a more robust option when creating real-world applications.

We study the motion prediction module because it is an indispensable task for planning safe and comfortable maneuvers. Recent works \cite{chai2019multipath, messaoudtrajectory} have highlighted two main factors that directly affect the agents future motion: The short term history of the agents movements and their interactions, and the scene context including road and crosswalk polygons, lane directions and boundaries, traffic lights, and other relevant map information. This task is specifically challenging due to the uncertainty of the future decisions of the agents, and it is seemingly intuitive predicting the future trajectories said agents is important.

We describe three models that utilize a Bird's Eye View (BEV) multi-channel input image representation that integrates both scene context, from a high definition map, and agents' motion history, obtained from a working object pose estimation module. All three models produce both multi-agent trajectory predictions and spatial uncertainty estimations. Our baseline model is the tracking free model. In order to evaluate the effect of the tracking module, we integrate the tracking information into two of the models. In one model, we integrate an Long-Short Term Memory (LSTM) embedding to represent the agent's past states based on its tracking information. In the second model, we integrate the identity information obtained from a tracking module in the BEV input image using displacement fields (to the best of our knowledge, this input representation is novel in the task of motion prediction). We evaluate the performance of our models under various conditions including synthetic tracking noise and using a real-world tracker and  conclude that, tracking is a sensitive step, if treated meticulously, can considerably improve the motion prediction performance. However, if it introduces noise to the system, it can degrade the model's performance and even become a hindrance to the system.
In summary, the main contributions of this paper are summarized in the following:
\begin{itemize}
    \item We conduct a comprehensive study to evaluate the effect of tracking noise on the motion prediction performance and compare the tracking-based to the tracking-free alternatives.
    \item We train three deep networks to predict short-term traffic agents trajectories and their spatial uncertainties in order to study the effectiveness of these modules under challenging conditions and evaluate the importance of tracking.
    \item We propose a novel input representation that integrates the tracking information using spatio-temporal displacement fields.
    \item We conduct various experiments to evaluate the performance of the three models under noisy tracking conditions, real-world tracking, and variable actor density conditions. 
    \item We conclude that, it is important to study the effect of tracking module on the motion prediction performance to avoid situations where the tracking module is a hindrance to the system’s performance.
\end{itemize}

%-------------------------------------------------------------------------
\section{Related Work}

Motion prediction has a long and storied history \cite{rasouli2020deep, yurtsever2020survey}, but, due to space constraints, we will limit this section to approaches that contextualize their efforts in the self-driving systems domain. We first cover classical approaches and then discuss deep learning based approaches by distinguishing the methods that use agent identification information from those that do not use such information.

Most of the deployed systems in industry use well-established, engineered approaches for motion prediction. One common approach is the Kalman Filter (KF) \cite{kalman1960new, wan2000unscented}. The KF estimates the agent's state and propagates it into the future based on kinematic models and assumptions of an underlying physical system. While the KF is efficient for short-term predictions, its performance degrades with longer term predictions because it is mainly agent-centric i.e., ignores external constraints (environment constraints, other agents ...) and only considers spatio-temporal agent information. Other classical approaches rely on machine learning models, such as Hidden Markov Models \cite{streubel2014prediction}, Gaussian Mixture Models \cite{yoo2016visual}, Processes \cite{mogelmose2015trajectory} and other techniques to solve this task. However, the real-world performance of these approaches usually suffers from high computation time, generalization issues (when confronted with noisy detections), or complex agent/environment interactions modeling issues.

Various deep learning approaches have been proposed to model agents behaviors in motion prediction task. Like other sequence prediction tasks, many motion prediction methods rely on recurrent architectures, such as LSTMs \cite{fragkiadaki2015learning,lee2017desire,  zhao2019multi}, to model the agents dynamics. Social-LSTM \cite{alahi2016social} uses LSTM embeddings to model each pedestrian's motion and then aggregates these embeddings using a social pooling technique to model inter-pedestrians interactions, before predicting their future trajectories. Social-GAN \cite{gupta2018social} extends Social LSTM by proposing a generative model based on Recurrent Neural Network (RNN). Giuliari \etal \cite{giuliari2020transformer} leverages Transformers (TFs), a recent technique developed within the NLP field for word sequences modeling, to model pedestrians trajectories. However, these methods have generalization issues when confronted with noisy detections \cite{rhinehart2019precog}.

Several model the interactions among agents using Graph Neural Networks. Social-BiGAT \cite{kosaraju2019social} relies on graph attention networks to model the social interactions between pedestrians where each node in the graph is a pedestrian represented using an LSTM embedding. Social-STGCNN \cite{mohamed2020social} directly models the pedestrian trajectories as a graph and uses a Time-Extrapolator Convolution Neural Network to operate on the temporal dimension. Though these methods capture the interactions among the agents, they fail to capture the environment constraints.
 %Furthermore, they only consider the interactions among agents whilst ignore the environmental context and its constraints.
VectorNet \cite{gao2020vectornet} suggests addressing this failing using polyline subgraphs to represent separate entities including the agents and the environment constraints. These subgraphs then form a global interaction graph that captures interactions among all environment components. Polyline representation allows graph-based approaches to account for the agents interactions with other environment components; however, it is often hard to extract automatically from sensors, and requires human annotations. 

In self-driving domain, a bird’s eye view (BEV) raster representation is widely used as input followed by Convolutional Neural Networks (CNNs)\cite{chai2019multipath, cui2019multimodal, djuric2020uncertainty,salzmann2020trajectron++}. The raster encodes the agents history information, context and other map information which allows the network to extract useful appearance features of the agents and their environmental context in order to predict their future trajectories. Djuric \etal \cite{djuric2020uncertainty} uses CNNs to predict short term vehicle trajectories from a BEV raster input, encoding individual agent's surrounding context. This work was later applied to Vulnerable Road Users (VRUs)\cite{chou2019predicting}. Chai \etal \cite{chai2019multipath} leverages a fixed set of future state-sequence anchors that correspond to modes of the trajectory distribution. 
Other approaches \cite{chandra2019traphic, djuric2020uncertainty, liang2020pnpnet} take advantage of both LSTMs and CNNs by proposing hybrid models that encode agents states using LSTMs and capture scene context and agents interactions using CNN features.

The existing methods we have discussed use engineered or learned techniques that capture the agent's past movements in order to predict their future trajectory. Some methods also consider the agent's interactions with other traffic actors and scene context to better forecast their future movements. A major assumption of these techniques is that the identity of the agents is known through time (i.e., agent tracking is perfect); however, in real-world applications, the performance of tracking will inevitably be imperfect, leading to identity switches \cite{chiu2020probabilistic, chaabane2021deft} (due to heavy occlusion, high inter-frame motion, crowded scenes, or high sensor motion \cite{caesar2020nuscenes}, to name but a few issues). Some approaches circumvent this issue with end-to-end methods that jointly learn detection and motion prediction directly from sensor data \cite{casas2018intentnet, djuric2020multinet, luo2018fast, sadat2020perceive,chaabane2020looking, trabelsi2021pose}, but they do not consider real-world error either. In this work, we explicitly study the interactions noise in the tracking module has on motion prediction performance by comparing models that integrate tracking information with models that do not use such information.

%-------------------------------------------------------------------------
\section{Methods}\label{methods}
\begin{figure*}
\begin{center}
% \fbox{\rule{0pt}{1in} \rule{0.9\linewidth}{0pt}}
   \includegraphics[width=0.8\linewidth]{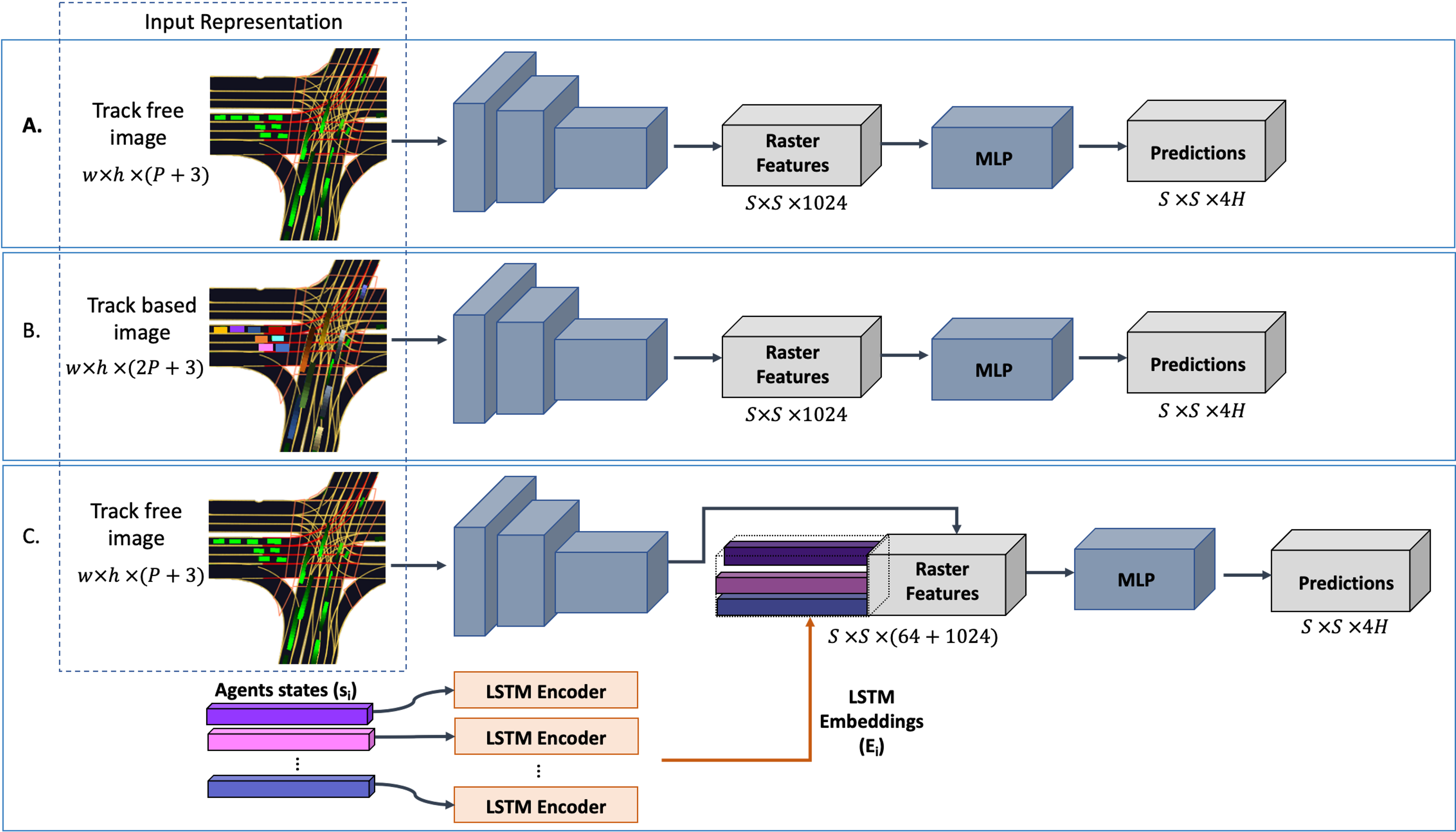}
\end{center}
   \caption{Overview of the three described architectures. We show the input representation rasterized into color-coded RGB image for visualization purposes. Each historical agent polygon is rasterized with the same color as the current polygon but with reduced level of brightness, resulting in the fading effect. \textbf{A.} represents the Track-free CNN method that relies on the tracking free input to predict the agents future trajectories. \textbf{B.} is the Track-based CNN method that integrates the tracking information in the input representation using displacement vector fields. \textbf{C.} shows the Hybrid method that extends \textbf{A.} by adding an LSTM encoding to represent each agent history. All agents in \textbf{A.} and \textbf{C.} inputs are represented with the same green color to show that no identity information was used to differentiate among agents. In \textbf{B.}, we represent each agent with different color to infer their identity information. The actual size of the input representation is $w\times h \times (nP+3)$ where $P$ is the number of past frames and $n$ is equal to 1 or 2 depending on the architecture.}
\label{architectures}
\end{figure*}

In this work, we analyse the importance of the tracking module for the task of motion prediction. 
To this end, we compare methods that integrate the identity information of agents to methods that remove the use of such information.
Here, we describe three methods (see \figurename~\ref{architectures}). The first is a fully convolutional model that outputs trajectory prediction of agents in the scene (\S~\ref{trackingfreeCNN}). The second is a hybrid CNN-LSTM model that extends the first model by integrating LSTM encoding extracted from agents' states (\S~\ref{hybrid}). The third is a novel fully convolutional model that integrates the identity information of agents in the BEV input image (further explained in \S~\ref{trackingbasedCNN}). All three models take multi-channel image input of a BEV of the scene and output multi-agent trajectory predictions.

For all models we assume to have access to a high definition map $\mathcal{M}$ of an operating area, comprising road and crosswalk polygons, lane directions and boundaries and other relevant map information. We assume models have a functioning pose estimation system ingesting the sensor data to detect and pose traffic actors.
Lastly, unless specified differently, we assume a perfect tracking system is available for the tracking-based models, providing ground-truth tracking of the detected traffic actors.
%-------------------------------------------------------------------------
\subsection{Input representation:}

We encode the static map elements from the high definition map $\mathcal{M}$ in a bird's eye view (BEV) image centered on the self-driving vehicle (SDV) where each element of the map, including driving lanes, crosswalks and traffic lights, is encoded as a binary mask in its own separate channel. These channels are then rasterized into an RGB image where each element is assigned a different color, as described in \cite{djuric2020uncertainty}.
Furthermore, we consider $P$ additional channels stacked with the map raster where each channel represents the agents locations at each timestep of the history and present. Each of these channels is a binary mask encoding the agents top down positions in the same BEV frame as introduced above. The final input is then formed of $P+3$ channels representing map information and agent's history and present locations. We note here that no identity information is inferred as all detections of agents at each timestep are treated similarly.
%-------------------------------------------------------------------------
\subsection{Tracking free CNN model:}\label{trackingfreeCNN}

In this model, we follow \cite{djuric2020multinet} and process the input multi-channel image using a sequence of 2-D convolutions to produce a dense feature representation for each grid cell of the input. We further add three $1\times1$ convolutional layers to finally output a 3-D tensor of size $S\times S\times 4H$ representing the predicted future movements of the agents present in the scene, where $S\times S$ is the size of the grid and $H$ is the number of future predictions. For each grid cell containing an agent center at the present timestep, we predict the 2-D centers offsets $(\Delta c_x, \Delta c_y)$ in $H$ future time horizons. In addition to predicting the future trajectories, we also estimate the spatial uncertainties of our predictions. %Similarly to \cite{djuric2020multinet}, we decompose the centers location uncertainty in the along-track (AT) and cross-track (CT) directions \cite{gong2004methodology} and assume the uncertainty in each direction follows an increasing linear function with time where the function parameters are model hyper-parameters (see \S~\ref{loss} for more details). 
Note that this model utilizes no prior identity information nor tracking step.
%-------------------------------------------------------------------------
\subsection{Hybrid model:}\label{hybrid}

The hybrid model is an extension of the first model (\S~\ref{trackingfreeCNN}) which further integrates an LSTM sequence model \cite{messaoudtrajectory}. The LSTM 
encodes each agent's states across past and present timesteps into a single embedding ($E_i$) for agent $i$. In our experiments, an agent state $s_t$ comprises position displacements with respect to the present, relative position changes, and speed at each timestep $t$ where $t \in {1,...,P}$ and $s_1$ represents the present state. For each agent, the LSTM embedding is concatenated with CNN features extracted from the CNN network (as in first model) at the grid cell containing the agent's center at the present timestep. The grid cells that do not contain agent centers are padded with zeros. The obtained feature block is then processed, similarly with the first model, with three $1\times 1$ convolutional layers and output a tensor of size $S \times S \times 4H$ representing the future trajectories and the corresponding uncertainties.

Note that the use of LSTM to encode an agent's past trajectory relies on the assumption that the identity of the agent is well known through time. %Comparing with the track-free CNN model  allows us to investigate how important the tracking information is and to what extent it improves the performance of motion prediction models. Further, this architecture also allows us to evaluate hybrid models and highlight the effect of the sequence-based component on the performance.
%-------------------------------------------------------------------------
\subsection{Tracking based CNN model:}\label{trackingbasedCNN}

This model follows the same architecture as the first model (\S~\ref{trackingfreeCNN}). The main difference resides in the input representation; in this model, we integrate the identity information in the input image. Specifically, instead of representing each timestep from the past with a binary mask to indicate the presence/absence of a detection at each pixel, we consider a spatio-temporal displacement vector field $D \in \mathbf{R}^{w\times h\times 2}$ at each timestep, where a 2-D vector at each pixel parallels the vector from the agent center at that timestep, to the center of the agent at the present timestep. $w$ and $h$ are the width and height of the input image. At the present timestep, we use a simple binary mask, similar to the initial input representation. The final input image then has $2P+3$ channels.

Like first model, this model relies on CNNs to operate on the spatial and the temporal dimension simultaneously and thus it is smaller in size compared to the hybrid model that uses both CNNs and LSTMs (\S~\ref{hybrid}).
Though displacement vector fields are a common representation in the segmentation task \cite{ahn2019weakly,neven2019instance}, to the best of our knowledge, we are the first to apply this technique in the motion prediction task.

%-------------------------------------------------------------------------
\subsection{Loss Function:}\label{loss}

In this paper, we train both trajectory prediction and uncertainty estimation jointly. We project the prediction errors on the along-track (AT) and cross-track (CT) directions using the ground-truth heading of agent, and we assume that each projected error in one of the two directions is independent from the other and follows a Laplace distribution $Laplace(\mu,b)$ with a PDF of a random Laplacian variable $v$ computed as:
\begin{equation}
\frac{1}{2b} \exp(-\frac{|v-\mu|}{b})
\end{equation}
where mean $\mu$ and diversity $b$ are the Laplace parameters.
Ideally the AT and CT errors would follow a ground-truth distributions of mean $\mu=0$ and diversities $b_{AT}$ and $b_{CT}$, respectively. Since, we assume that the uncertainty increases with time. We define the diversity as a linearly increasing function:
\begin{equation}
b_i = \alpha_i t + \beta_i
\end{equation}
where $\alpha_i$ and $\beta_i$ are model hyper-parameters defined separately for AT and CT.
To train the model, we minimize the Kullback-Leibler (KL) divergence between the ground-truth distribution $Laplace(0, b_i)$ and the predicted distribution $Laplace(\hat{e}_i, \hat{b}_i)$ as in \cite{djuric2020multinet} defined as:
\begin{equation}
KL_i = \log(\frac{\hat{b}_i }{b_i })+ \frac{b_i  \exp(-\frac{|\hat{e}_i |}{b_i })+|\hat{e}_i |}{\hat{b}_i } - 1
\end{equation}
 where $i$ is whether AT or CT. 
%--------------------------------------------------------------------------
\section{Experimental Details:} \label{experiments}

In this section, we describe the dataset we used to run the different experiments (\S~\ref{dataset}). We also give details of the architectures and experimental settings (\S~\ref{settings}).
%-------------------------------------------------------------------------
\subsection{Dataset:} \label{dataset}
The goal of this work is to analyse different aspects of motion prediction models and not to compare the performance of our models with state-of-the-art methods. To this end, we use the Lyft Prediction Dataset \cite{houston2020one} to run our experiments. The Lyft Prediction Dataset is the largest public self-driving dataset for motion prediction to date, with 1,118 hours of recorded self-driving perception data. It was collected by a fleet of 20 autonomous vehicles along a fixed route in Palo Alto, California over a four-month period. It consists of 170,000 scenes, 25 seconds long each capturing the positions and motions of the surrounding agents including vehicles, cyclists and pedestrians. The dataset also comprises a high-definition semantic map with 15,242 labelled elements and a high-definition aerial view over the area.

%-------------------------------------------------------------------------
\subsection{Experimental settings:}\label{settings}

For input representation, we use a BEV image with spatial horizontal dimensions $512 \times 512$, where each grid cell is $0.25m \times 0.25m$. For temporal information, we consider a history of 1s (equivalent to 10 timestamps at 10Hz) resulting in an input of size of $w \times h \times nP + 3$ where $n$ is the number of channels per timestamp ($n=1$ for both tracking-free CNN model and hybrid model and $n=2$ for tracking based CNN model) and $P=11$ (10 past frames and 1 present frame). We chose to use 1s of history for real-time efficiency following \cite{djuric2020multinet, liang2020pnpnet}. The output tensor is of size $128 \times 128 \times d$ where $d = 4H$ channels.
For the backbone network, we use ResNet-50 \cite{he2016deep} to extract deep features of size $S \times S \times 1024$. The models were implemented in PyTorch \cite{paszke2019pytorch} and trained from scratch with a batch size of 4 for 2 epochs with Adam optimizer \cite{kingma2014adam}, setting the initial learning rate to $10^{-4}$ that was further decreased by a factor of $0.9$ every $200$ thousand iterations. We ran our experiments on a Ubuntu server with a TITAN X GPU with 12 GB of memory. For our experiments, we report along-track (AT) error metric and cross-track (CT) error metric \cite{gong2004methodology}, as well as the average displacement error (ADE) and the final displacement error (FDE) \cite{alahi2016social}. All metrics are reported on the validation dataset as specified in \cite{houston2020one} at an horizon of $H = 50$ equivalent to 5 seconds in the future.

%------------------------------------------------------------------------
\section{Results}\label{results}
It is a common practice in the field of motion prediction to rely on the agent's past sequence of tracks in order to predict their future trajectory. Such practice makes a major assumption on the availability of a robust tracking system that provides little-to-no-noise identity information to the agents in the scene. Such assumption does not always hold true as the tracking system is always prone to noise due to several factors including occlusion, crowded scenes, high inter-frame motion... 
In this section, we first (\S~\ref{overall}) evaluate the performance of the motion prediction methods described in \S~\ref{methods} and compare models that integrate the identity information of agents to models that do not use such information. We further (\S~\ref{density_evaluation}) depict these results by considering scenarios where the knowledge of the agent identity may play a crucial role in the performance of the model prediction such as the case of crowded scenes.
We also evaluate the effect of noise in tracking on the performance of the models by applying synthetic noise (\S~\ref{noisy_tracker}) as well as realistic noise coming from real-world tracker (\S~\ref{real_tracker}). 

%-------------------------------------------------------------------------
\subsection{Overall Performance Evaluation}\label{overall}

\begin{table}
\small
    \begin{center}
        \begin{tabular}{|l|c|c|c|c|}
        \hline
        Method & AT & CT & ADE & FDE\\
        \hline\hline
        Track-free CNN & 1.241 & 0.571 & 1.379 & 2.577 \\
        Track-based CNN & 1.232  & \textbf{0.549} & \textbf{1.328} & 2.556 \\
        Hybrid & \textbf{1.229} & 0.567 & 1.345 & \textbf{2.552}  \\
        \hline
        \end{tabular}
    \end{center}
    \caption{Overall comparison of the described methods using four metrics in meters. Given a noise free tracking system, the Hybrid model performs the best in AT and FDE metrics, while the Track-based model performs the best in CT and ADE metrics.}
    \label{overallrslts}
\end{table}

Results of the three models are summarized in \tablename~\ref{overallrslts} with best prediction results shown in bold. We compare the performance using 4 different metrics AT, CT, ADE and FDE (as introduced in \S~\ref{settings}) averaged over a prediction horizon 5s. We note that even small metric improvements can make a significant difference in the performance and safety of the real-world system. 

\begin{figure*}
\begin{center}
% \fbox{\rule{0pt}{1in} \rule{0.9\linewidth}{0pt}}
   \includegraphics[width=\linewidth]{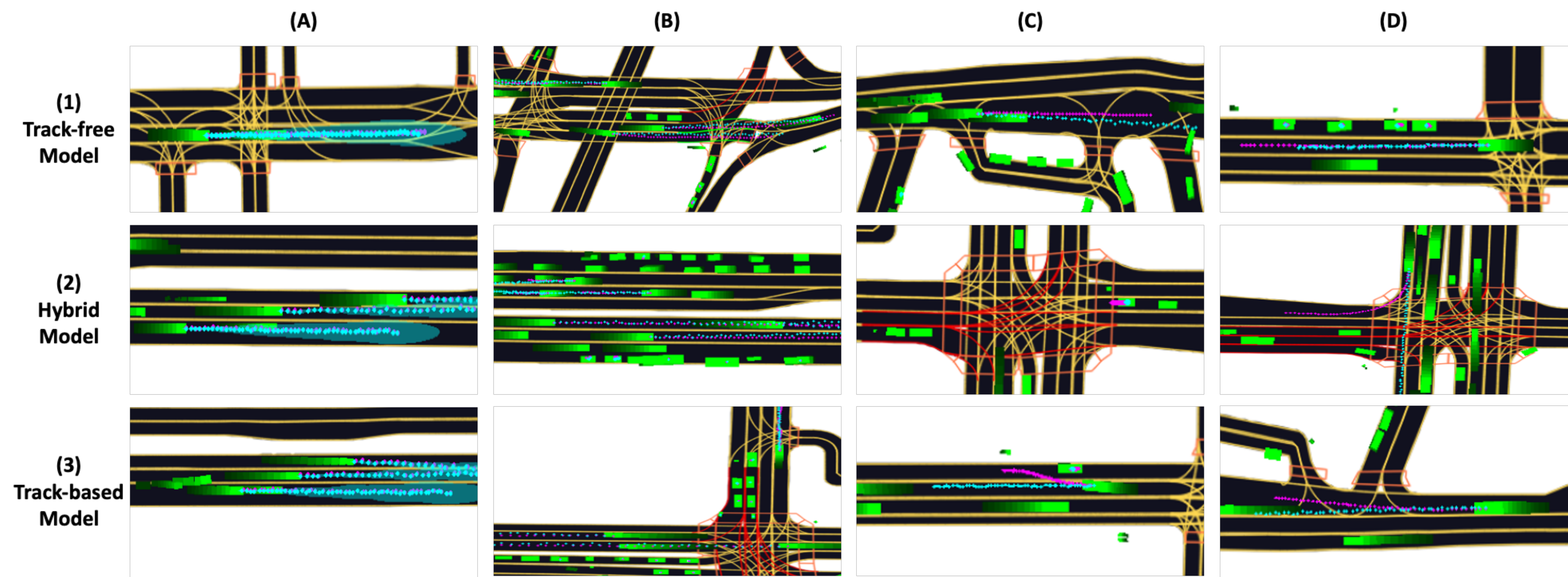}
\end{center}
   \caption{Results examples of the described methods. We plot the target trajectories in Magenta and the predicted trajectories in Cyan. For clearer visualization, The scenes are zoomed in and only the trajectories of a subset of the agents are shown. Rows \textbf{(1)}, \textbf{(2)}, and \textbf{(3)} show examples using the Track-free CNN model, Hybrid model and Track-based CNN model respectively. Columns \textbf{(A)} and \textbf{(B)} show success cases. In \textbf{(A)}, we also plot the uncertainties of the predicted trajectories in light Cyan. Columns \textbf{(C)} and \textbf{(D)} show failure cases. Examples (1)-(C), (2)-(D), (3)-(C) and (3)-(D) show failure in the estimation of the future direction of the agents. We see high error in the cross-track direction. (1)-(C), (1)-(D), (2)-(C) and (3)-(C) show failure in the estimation of the velocity of the agents. Thus we see high error in the along-track direction.}
\label{qualitative}
\end{figure*}

Using the tracking information in both the Track-based CNN and Hybrid models improve the performance by 3\% and 2.5\% respectively, compared to the Track-free CNN model. Comparing the Track-based CNN and the Hybrid, the latter obtains better prediction accuracy on the AT and FDE metrics. This is unsurprising, as LSTMs are efficient in learning long-term temporal dependencies and thus can better capture the agent dynamics such as velocity and acceleration. Furthermore, the Track-based CNN model performs better than the Hybrid model in terms of CT and FDE. Thus, such model would perform better in lane association or in passing scenarios.
In \figurename~\ref{qualitative} we show qualitative examples of 2 success cases and 2 failure cases for each of the three models described in this work.

%-------------------------------------------------------------------------
\subsection{Model Performance Depends on Agent Velocity and Traffic Density}
\label{density_evaluation}
Since the dataset encloses a variety of scenarios with large amounts of behavioural observations and interactions, it is hard to depict the effect of the tracking module by evaluating the full testing data. Based on preliminary studies, we found that all three models perform equally well in the scenarios where agents are moving slowly or are stationary. We then categorize the scenes using the agent's velocity and the density of the scene and limit our future comparisons to scenes with agent velocities $v$ larger than $3m/s$.  Furthermore, we categorize these agents based on the density of their surrounding environment. We measure the density by calculating the distance between the agent and their nearest neighbour. We consider a scenario "dense" if the agent's distance is less than 4 meters ($r < 4$) and a scenario "non dense" if the distance is larger than 10 meters ($r > 10$). 

\begin{figure}[t]
\begin{center}
% \fbox{\rule{0pt}{1in} \rule{0.9\linewidth}{0pt}}
   \includegraphics[width=0.7\linewidth]{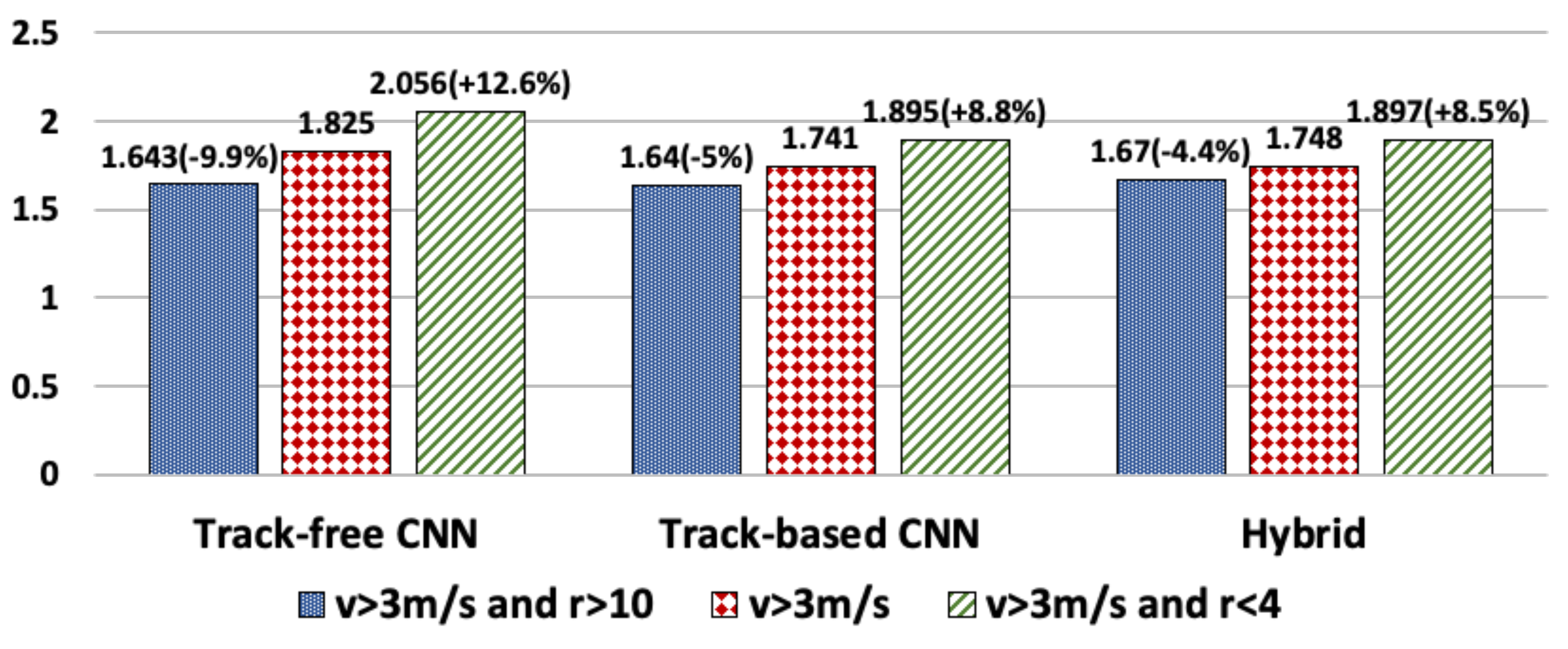}
\end{center}
   \caption{Performance evaluation of the described methods in ADE (m) on agents moving with a velocity larger than $3m/s$. We further consider the scene density factor. We calculate the distance $r$ between a given agent and their closest neighbor and select those with $r<4m$ in one experiment (dense scenarios) and $r>10m$ in a second experiment (non dense scenarios). The performance of the three methods degrade in dense scenarios. The decrease is most pronounced with Track-free model which shows the importance of tracking information under these conditions.}
\label{density}
\end{figure}

The results are summarized in \figurename~\ref{density}. We show a bar plot with the performance of each scenario in ADE metric for each model. For dense and non-dense scenarios we specify the performance change (in percentages) with respect to all agents with $v>3m/s$. First, as expected, the performance decreased in all three models since the selected scenarios are relatively challenging due to the high inter-frame motion and density of the scenes. Second, the track-free model significantly decreases in the performance compared to the tracking based models. Comparing to all moving agents ($v>3m/s$), the performance of the track-free CNN model has decreased by 12.6\% on dense scenarios ($r<4$) compared to a more attenuated decrease of 8.8\% and 8.5\% for track-based CNN and hybrid models, respectively. On non dense scenarios ($r>10$), the three models perform more comparably. These findings rightly show that the tracking information can be essential in challenging scenarios, such as a very crowded scene where the input representation of the agents can become less effective. However, for other scenarios, such as non dense scenarios, the three models seem to perform comparably well. This is the noise-free condition --- in Sections \ref{noisy_tracker} and \ref{real_tracker}, we reevaluated our three models in the context of tracking noise.

%-------------------------------------------------------------------------
\subsection{Performance Evaluation with Noisy Tracker} \label{noisy_tracker}

\begin{figure}[t]
\begin{center}
% \fbox{\rule{0pt}{1in} \rule{0.9\linewidth}{0pt}}
   \includegraphics[width=0.7\linewidth]{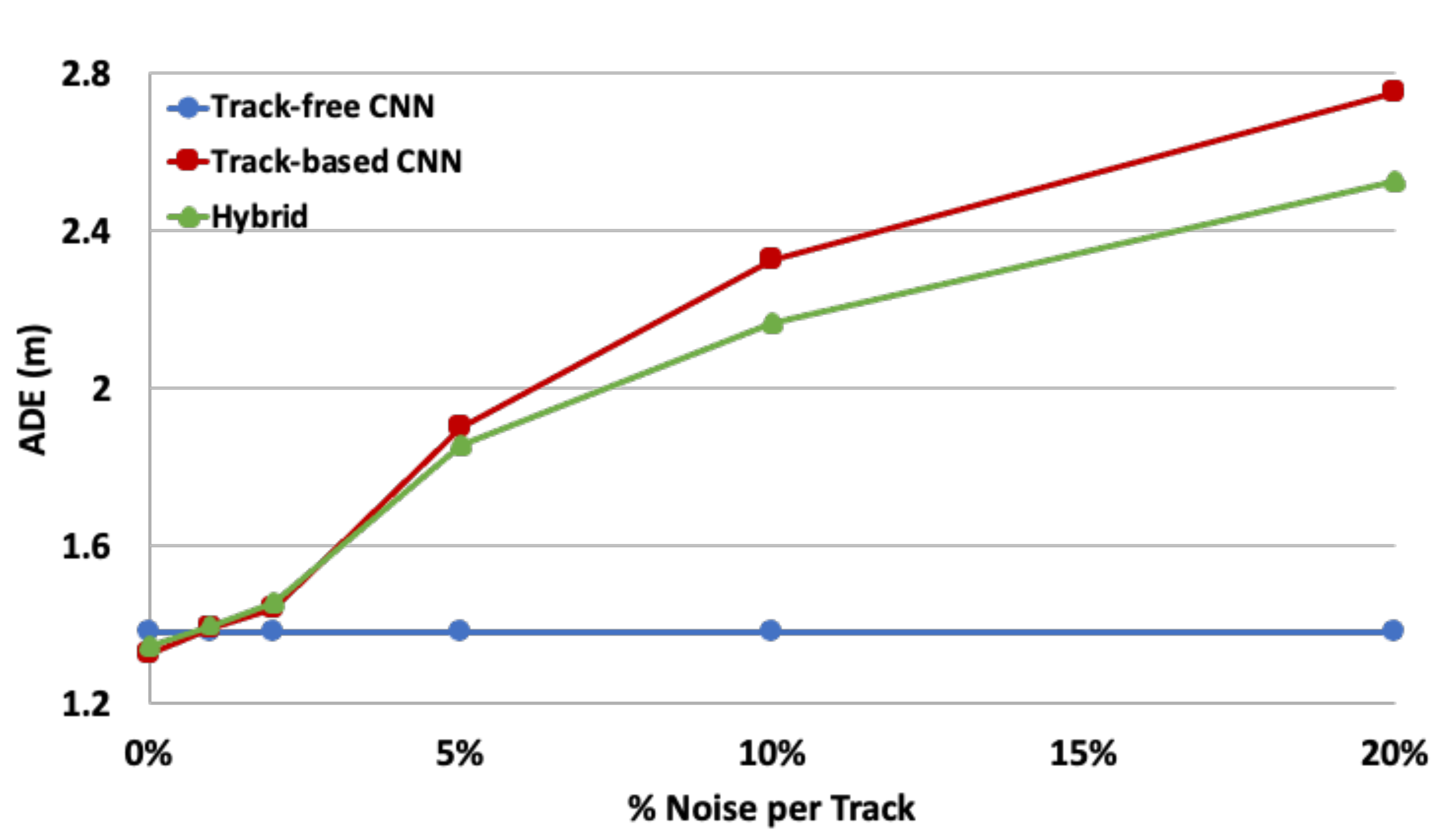}
\end{center}
   \caption{Performance Evaluation of the described methods in ADE (m). We apply synthetic noise to the tracking information. We experiment with an increasing chance of 1 identity switch per track. The performance of the tracking based methods (track-based CNN and Hybrid) decrease with increasing tracking noise. The track-free model is not affected by tracking noise.}
\label{noise1}
\end{figure}

\begin{figure}[t]
\begin{center}
% \fbox{\rule{0pt}{1in} \rule{0.9\linewidth}{0pt}}
   \includegraphics[width=0.8\linewidth]{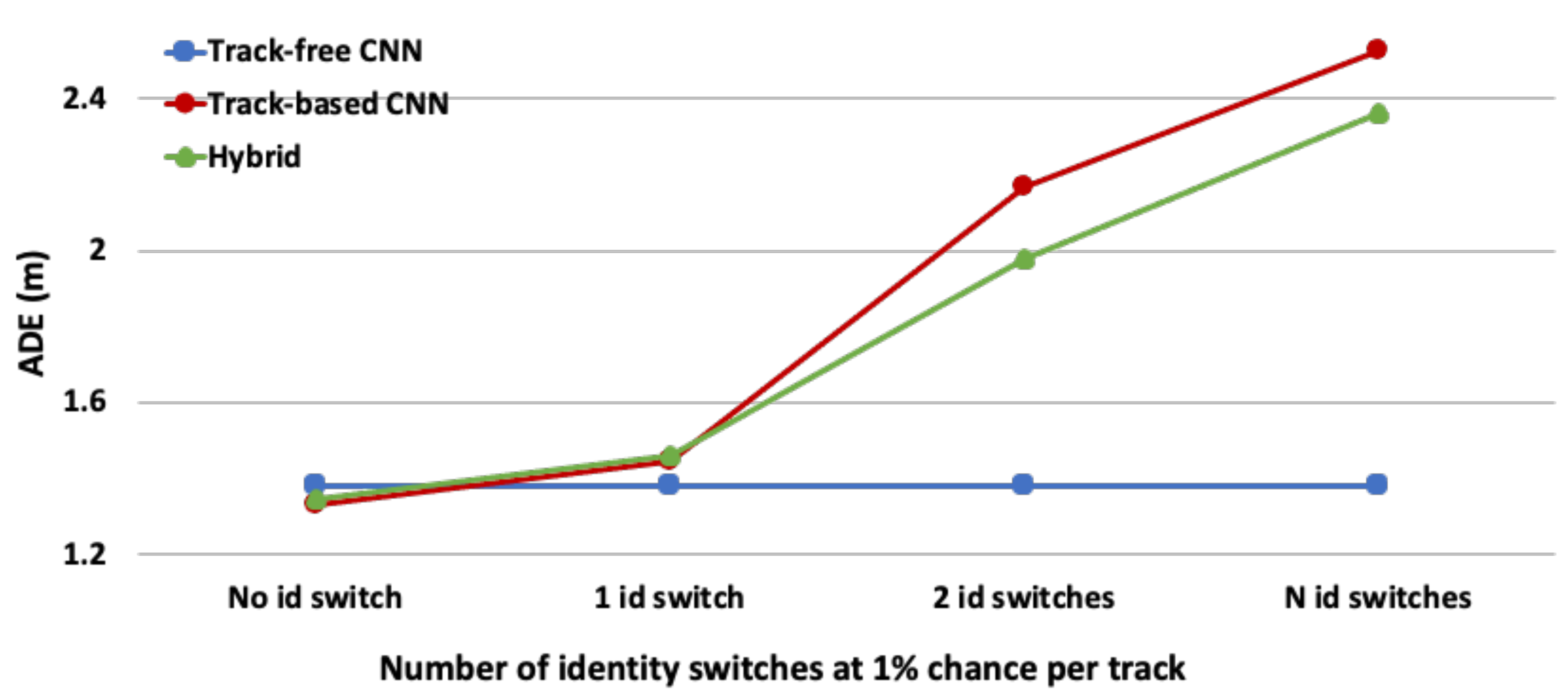}
\end{center}
   \caption{Performance Evaluation of the three methods in ADE (m) with synthetic tracking noise of 1\% chance. We experiment with an increasing number of identity switches per track. 1 id switch represent identity switch at only 1 timestamp. 2 id switches represent an identity switch for 2 consecutive timestamps and N id switch represent 1 identity switch that started at a random timestamp and continued until the end of the scene. The performance of the tracking based methods (track-based CNN and Hybrid) decrease with increasing tracking noise. The track-free model is not affected by tracking noise.}
\label{noise2}
\end{figure}

In these experiments, we evaluate the effect of noisy tracker on the performance of the three models. Being independent from the tracking system, the performance of the track-free CNN model remains constant in these experiments. We apply synthetic noise to the tracking system and evaluate its effect on the performance of the two tracking based models. In the first set of experiments, shown in \figurename~\ref{noise1}, we perform random identity switches with varying chances per track. We vary the probability of an identity switch from 0\% to 20\% per track. The performance of both the track-based CNN and hybrid model has decreased with increasing noise, degrading slightly around 1\% chance and then drastically after 2\% chance of identity switch per track. Comparing the tracking based models to the track-free CNN model, the latter obtains better performance on all experiments with noise larger than 0.8\%. The drastic decrease in the performance of both the track-based CNN and Hybrid model shows that they both rely heavily on the tracking information to capture the agents' past movements and thus at a certain level (around 0.8\% noise chance), the cascaded tracking noise starts to negatively affect the performance of these models. With noise chance larger than 1\%, the tracking noise has a significant negative effect on the performance of these models. For experiments with noise chance larger than 2\%, the Hybrid model is clearly shown to be more robust than the Track-based CNN model. Though the Hybrid model is highly dependent on the LSTM input enclosing the tracking information, it also relies on the input image which is independent from the tracking module, which explains its relative robustness to noise compared to track-based CNN model.

The second set of experiments shown in \figurename~\ref{noise2}, comprises common identity switch scenarios. We set the identity switch chance to 1\% per track throughout the experiments and we consider three scenarios: an identity switch happening at a single random timestamp (1 id switch), an identity switch happening for two consecutive timestamps (2 id switches), and an identity switch happening at a random timestamp and continuing until the end of the scene (N id switches). Similarly to the first set of experiments, the performance of both the track-based CNN and hybrid model declines in all three scenarios and falls behind the performance of the track-free model. The Hybrid model, for instance, has decreased from 1.345 to 1.485 in ADE when applying 1 id switch with a 1\% chance. It then falls by 35.5\% and 59.1\% when applying 2 id switches and N id switches respectively. Similar to the findings of the previous experiments, the Hybrid model is more robust to noise compared to Track-based CNN model when dealing with 2 id switches and N id switches.

%------------------------------------------------------------------------
\subsection{Peformance Evaluation with Real-world Tracker:}\label{real_tracker}

\begin{table}
\footnotesize
\begin{center}
\begin{tabular}{|l|c|c|c|c|}
\hline
Method & AT & CT & ADE & FDE\\
\hline
Track-free CNN &  \textbf{1.241} & \textbf{0.571} & \textbf{1.379} & \textbf{2.577}\\
Track-based CNN &  1.268& 0.607& 1.478& 3.013\\
Hybrid & 1.263 & 0.611 & 1.485 & 2.987 \\
\hline
\end{tabular}
\end{center}
\caption{Overall comparison of the described methods using four metrics (m) using StanfordIPRL-TRI tracker \cite{chiu2020probabilistic}. Real-world trackers introduce noise to the tracking information which affects the performance of track-based methods (Track-based CNN and Hybrid methods).}
\label{real_tracker_rslts}
\end{table}

\begin{figure}[t]
\begin{center}
% \fbox{\rule{0pt}{1in} \rule{0.9\linewidth}{0pt}}
   \includegraphics[width=0.7\linewidth]{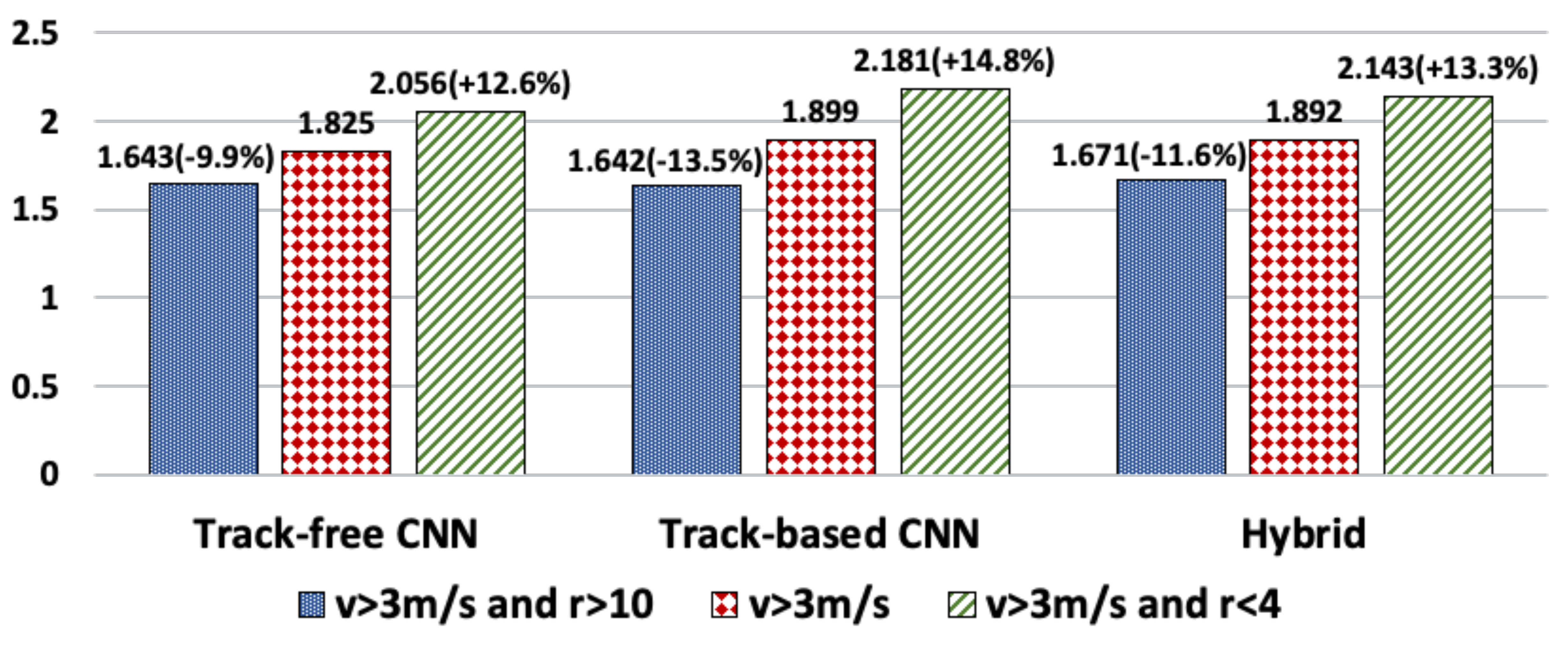}
\end{center}
   \caption{Performance evaluation of the described methods in ADE (m) using StanfordIPRL-TRI tracker \cite{chiu2020probabilistic} on agents moving with a velocity larger than 3m/s and different scene density. The performance decrease of the track-based methods is more pronounced using the real-world tracker due to the noise introduced to the tracking information.}
\label{density_realtracker}
\end{figure}
In this section we evaluate the performance of the tracking based models using a real-world tracker. To this end, we replace the tracking information provided in the dataset with the output of a real-world, popular tracker and reproduce the experiments that we have conducted in the sections \S~\ref{overall} and \S~\ref{density_evaluation} on the tracking based models. We use the StanfordIPRL-TRI tracker introduced in Chiu \etal \cite{chiu2020probabilistic} to run our experiments. We have utilized their publicly available code to run our experiments and selected the same parameters as suggested in their work. The StanfordIPRL-TRI tracker won the nuscenes challenge competition \cite{caesar2020nuscenes} by achieving state-of-the-art results on the nuscenes dataset.

The results of the Track-based CNN and Hybrid models using the StanfordIPRL-TRI tracker are summarized in \tablename~\ref{real_tracker_rslts}. The track-free CNN model does not depend on the tracking module so its performance remains the same as in \tablename~\ref{overallrslts}. The results suggest that there is a slight drop in the performance of the two tracking based models when using the StanfordIPRL-TRI tracker. Compared to the track-free CNN model, the track-based CNN model falls slightly behind by 2.17\% in AT and 6.3\% in CT. Similarly, the Hybrid model drops back by 1.77\% in AT and 7\% in CT with respect to the track-free CNN model. These findings suggest that state-of-the-art trackers can introduce noise that will be cascaded to the motion prediction module. If practitioners do not take preventive measures, the introduced noise can ultimately affect the motion prediction performance. These results also show that, in the case of noisy tracking information, the tracking free model performs better than the tracking based models which makes it a potential alternative to avoid cascaded noise. Alternatively, the motion prediction model can be trained to overcome noisy inputs from the tracking step and thus robustly recovers accurate predictions even under challenging conditions. 

Further experiments are conducted to evaluate the performance of models using the StanfordIPRL-TRI tracker on challenging scenarios, as described in \S~\ref{density_evaluation}, where we select the agents that moved at a speed higher than $3m/s$. We also consider the dense-versus-non-dense scenarios where the closest neighbor to the agent was located at a distance less than 4 meters and larger than 10 meters, for dense and non dense respectively. Results of this experiment are outlined in \figurename~\ref{density_realtracker}.
We notice that, similarly to the track-free CNN model, the tracking based models performance has dropped. The performance of the Track-based CNN model and Hybrid model drops by 4\% and 3.6\%, respectively, compared to track-free CNN model. For the dense scenarios, the track-based CNN performance drops by 14.8\% compared to the ``all moving agents scenario" ($v>3m/s$) (\S~\ref{density_evaluation}), while the track-free model has dropped by 12.6\%. These results suggest that though the performance has decreased across the three models due to the complexity of the scenarios, the track-free model performs the best among the three models. Furthermore, the decrease in performance is more highlighted in the more challenging scenarios ($v>3m/s$ and $r<4$) where the Track-based CNN model's ADE decreases to 2.181m as opposed to 2.056m for the track-free CNN model. This finding is unsurprising since the tracking based models are more affected by the tracking noise in more challenging scenarios where they heavily rely on tracking information.

\begin{figure}
\begin{center}
% \fbox{\rule{0pt}{1in} \rule{0.9\linewidth}{0pt}}
   \includegraphics[width=\linewidth]{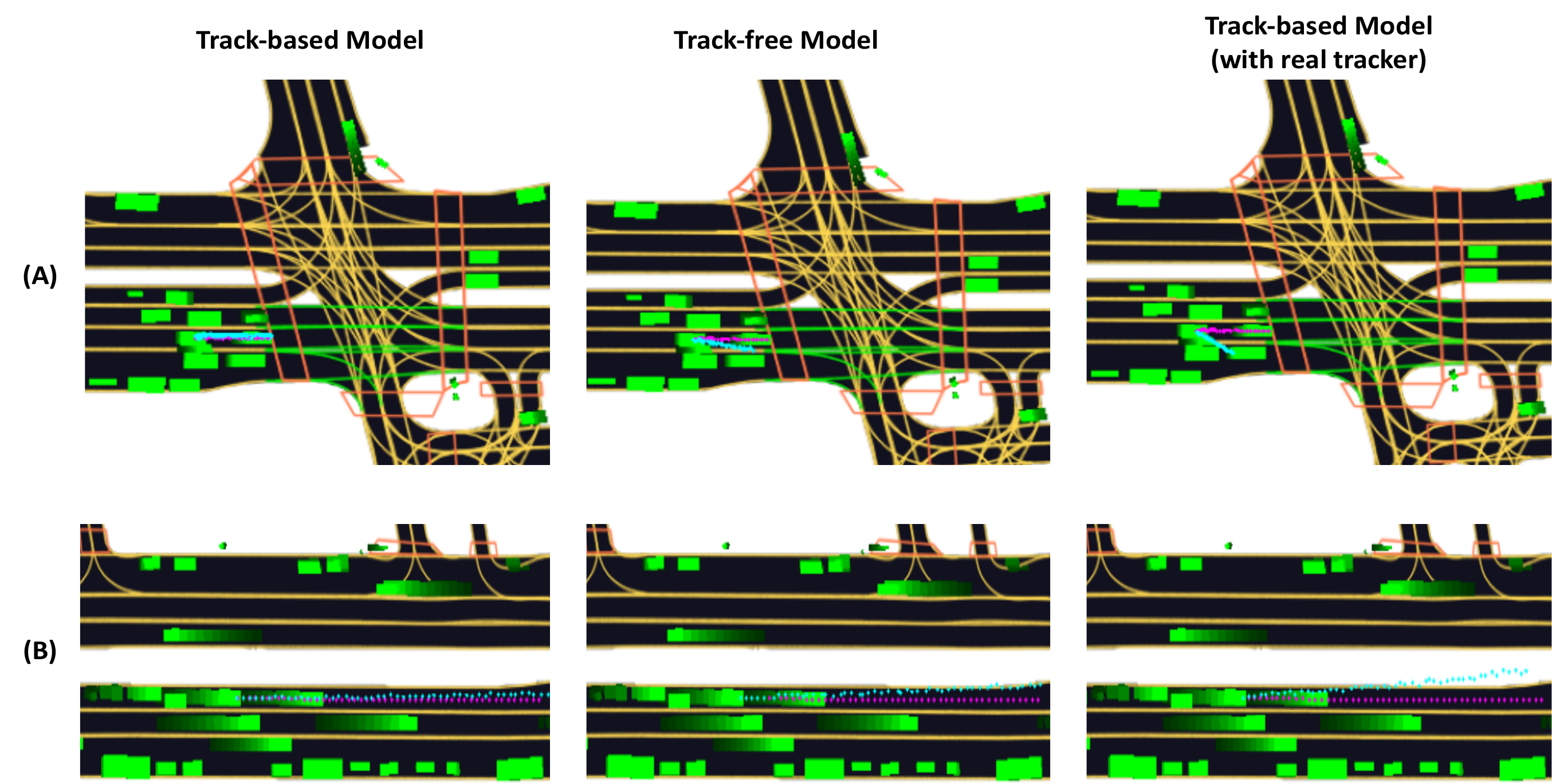}
\end{center}
   \caption{Qualitative evaluation in the case of a crowded scene where an identity switch happened. We show the performance of the track-based CNN model using ground-truth tracking \textbf{(1)}, the track-free CNN model \textbf{(2)}, and the track-based CNN model using StanfordIPRL-TRI tracker \textbf{(3)}, in 2 examples \textbf{(A)} and \textbf{(B)}.}
\label{qualitative_tracking}
\end{figure}

In \figurename~\ref{qualitative_tracking} we highlight examples of challenging scenarios, with crowded scenes where identity switches happened, and compare the performance of the Track-based CNN model, using real-world tracker \cite{chiu2020probabilistic}, with the Track-free CNN model and Track-based CNN model using the ground-truth tracking information (no identity switch for this model). Comparing the first and third row, we see that the performance of the track-based model, using the real-world tracker, (third row) degrades compared to the track-based model using the ground-truth tracker (first row) in the presence of identity switches. Comparing the second and third row, both models do not perform well in the two proposed scenarios. However, the track-free model is more robust to crowded scenes.
%------------------------------------------------------------------------
\section{Conclusion}

We proposed a comprehensive evaluation of three motion prediction models. The first, the Track-free CNN model, operated on a BEV input that was created based on a high definition map and agents detections, with no tracking information included. The second, the Hybrid model, extended the first model by integrating the tracking information in the form of LSTM embeddings of the agents past states. The third, the Track-based CNN model, integrated the tracking information in the BEV input using spatio-temporal displacement fields. We experimentally compared our models across no-noise, synthetic noise, and real-noise conditions and under various conditions. 

Results indicated that, while the tracking-based models performed better than the track-free model in the noise-free condition, their performance rapidly degraded when the tracking system produced noise --- resulting in them falling behind the tracking-free system performance. The takeaway from this study is that practitioners should be aware of the effect of tracking noise on the motion prediction performance and should consider it when selecting their approach. Furthermore, the tracking-free models can be, in the case of an inevitable noisy tracker, a potential option that is more robust when creating real-world applications. Potentially, preliminary experiments could be conducted before deciding on the final approach. For instance, a comparative study on the motion prediction models with and without tracking step can be done. Alternatively, an end-to-end approach could be adopted with a thorough performance analysis to make sure that the motion prediction model learns to ignore noisy inputs from the tracker and robustly recovers accurate predictions under various conditions.

\bibliographystyle{unsrt}  
\bibliography{references}  %%% Remove comment to use the external .bib file (using bibtex).


% George Kour and Raid Saabne.
% \newblock Real-time segmentation of on-line handwritten arabic script.
% \newblock In {\em Frontiers in Handwriting Recognition (ICFHR), 2014 14th
%   International Conference on}, pages 417--422. IEEE, 2014.

% George Kour and Raid Saabne.
% \newblock Fast classification of handwritten on-line arabic characters.
% \newblock In {\em Soft Computing and Pattern Recognition (SoCPaR), 2014 6th
%   International Conference of}, pages 312--318. IEEE, 2014.

% \bibitem{hadash2018estimate}
% Guy Hadash, Einat Kermany, Boaz Carmeli, Ofer Lavi, George Kour, and Alon
% \newblock Estimate and replace: A novel approach to integrating deep neural
%   networks with existing applications.
% \newblock {\em arXiv preprint arXiv:1804.09028}, 2018.

% \end{thebibliography}

\end{document}